%% file: main.tex
\RequirePackage{amsmath}
\documentclass[runningheads]{llncs}
\usepackage{graphicx} 
\usepackage{amsmath}
\usepackage{ifthen}
\usepackage{adjustbox}
\usepackage{xkeyval}
\usepackage{tikz}
\usepackage{calc}
\usepackage{amssymb}
\usepackage{multirow}
\usepackage{paralist}
\usepackage{comment}
\usepackage[utf8]{inputenc} 
\usepackage[T1]{fontenc}    
\usepackage{url}            
\usepackage{booktabs}       
\usepackage[dvipsnames,svgnames]{xcolor}         
\usepackage{arydshln}
\usepackage{wrapfig}
\usepackage{bm}
\usepackage{xspace}
\usepackage{nicefrac}       
\usepackage{microtype}      
\usepackage[hidelinks]{hyperref}
\usepackage{amsmath,amssymb} 
\usepackage[normalem]{ulem}
\useunder{\uline}{\ul}{}
\usepackage{colortbl}
\usepackage[normalem]{ulem}
\usepackage{pifont}

\def\eg{\emph{e}.\emph{g}.}
\def\ie{\emph{i}.\emph{e}.}

\usepackage[capitalize]{cleveref}
\crefname{section}{Sec.}{Secs.}
\Crefname{section}{Section}{Sections}
\Crefname{table}{Table}{Tables}
\crefname{table}{Tab.}{Tabs.}

\newcommand{\mysection}[1]{\vspace{-0.0mm}\section{#1}\vspace{-0.0mm}}
\newcommand{\mysubsection}[1]{\vspace{-0.0mm}\subsection{#1}\vspace{-0.0mm}}

\newcommand{\similarity}[1]{\langle#1\rangle}

\begin{document}

\title{Geometry-Guided Local Alignment for Multi-View Visual Language Pre-Training in Mammography}

\author{
Yuexi~Du\inst{1},~
Lihui~Chen\inst{1},~
Nicha~C.~Dvornek\inst{1,2} 
}
\institute{Department of Biomedical Engineering, 
\and Department of Radiology \& Biomedical Imaging, \\Yale University, New Haven, CT, USA \\
\email{\{yuexi.du, leon.chen, nicha.dvornek\}@yale.edu}
}

\authorrunning{Y. Du et al.}

\titlerunning{Geometry-Guided Local Alignment for Multi-View VLP in Mammography}

\maketitle

\begin{abstract}
    Mammography screening is an essential tool for early detection of breast cancer. The speed and accuracy of mammography interpretation has the potential to be improved with deep learning methods. However, the development of a foundation visual language model (VLM) is hindered by limited data and domain differences between natural and medical images. Existing mammography VLMs, adapted from natural images, often ignore domain-specific characteristics, such as multi-view relationships in mammography. Unlike radiologists who analyze both views together to process ipsilateral correspondence, current methods treat them as independent images or do not properly model the multi-view correspondence learning, losing critical geometric context and resulting in suboptimal prediction.
    We propose \textbf{GLAM}: \textbf{G}lobal and \textbf{L}ocal \textbf{A}lignment for \textbf{M}ulti-view mammography for VLM pretraining using geometry guidance.
    By leveraging the prior knowledge about the multi-view imaging process of mammograms, our model learns local cross-view alignments and fine-grained local features through joint global and local, visual-visual, and visual-language contrastive learning. Pretrained on EMBED~\cite{jeong2023emory}, one of the largest open mammography datasets, our model outperforms baselines across multiple datasets under different settings. \footnote{The code is available at \url{https://github.com/XYPB/GLAM}.}
\end{abstract}

\input{parts/1_intro}

\input{parts/2_methods}

\input{parts/3_experiment}

\input{parts/4_conclusion}

\bibliographystyle{splncs04}
\bibliography{clip}
\setcounter{tocdepth}{1}

\end{document}

%% file: parts/1_intro.tex
\mysection{Introduction}

\input{floats/fig_teaser}

Mammography screening is an effective tool for early detection of breast cancer, one of the most deadly cancers~\cite{siegel2014cancer,sung2021global}. Unlike many natural or medical images that offer a single view, standard mammography protocol produces two 2D images of the same 3D breast from different angles -- craniocaudal (CC) and mediolateral oblique (MLO) (\cref{fig:teaser}(a)). This dual-view nature, known as \emph{ipsilateral correspondence}, requires special consideration in clinical interpretation. Radiologists rely on both views to accurately locate regions of interest (ROIs), such as tumors or calcifications, and to mitigate ambiguities caused by projection angles~\cite{jain2024follow,liu2021act,ji2023mammo}. For instance, as in \cref{fig:teaser}(b), an ROI (red dot) might lie anywhere along the vertical blue “tube”, resulting in the same CC view image, while their MLO appearance will be different due to the MLO imaging angle. Thus, ignoring either view can lead to diagnostic errors, especially in data-driven deep-learning models that lack prior knowledge of the imaging process.
In addition to considering prior imaging knowledge, inclusion of multi-modal information through contrastive language-image pre-training (CLIP)~\cite{radford2021learning} has shown promise in enhancing medical image analysis. However, most prior CLIP models in the medical domain focus on other modalities like chest X-ray~\cite{wang2022multi,zhang2022contrastive,wang2022medclip,wu2023medklip}. Meanwhile, mammography-specific models only conduct global alignment, neglecting fine-grained multi-view local alignment~\cite{chen2024mammo,ghosh2024mammo,du2024multi}. Besides, existing image-only multi-view mammography methods primarily use global feature fusion~\cite{akselrod2019predicting,chen2022multi,xia2023neural,sun2022transformer,jouirou2019multi,manigrasso2025mammography,chen2024braixdet}, which compromises local detail. Others consider local multi-view alignment, \eg, using graph neural networks to learn the cross-view attention \cite{liu2021act,liu2020cross} or feature cosine similarity to model the multi-view relationship \cite{jain2024follow}; however, they lack the geometry knowledge needed for correct alignment that follows the actual 3D breast structure.

In this paper, we propose \textbf{G}lobal and \textbf{L}ocal \textbf{A}lignment for the \textbf{M}ulti-view mammography CLIP foundation model with geometry guidance, \ie, \textbf{GLAM}.
Pre-trained on $\sim200k$ screening mammograms, our model is among the largest in this domain. Inspired by the mammography imaging process and geometry-guided patch matching~\cite{engeland2006finding,sahiner2006joint}, we propose a self-supervised, cross-view local patch alignment method that respects the CC and MLO projection relationship. 
Instead of patch-to-patch alignment that improperly treats the breast as a rigid body~\cite{liu2021act,yang2021momminet,jain2024follow,ma2021cross}, we adopt patch-to-slice alignment along the anterior-posterior (AP) axis (\cref{fig:teaser}(b)). We use cross-attention to include all relevant tissues along the AP slice while accounting for breast deformation in the CC-mediolateral (ML) plane during imaging. We evaluate our method on three datasets with varied distributions~\cite{jeong2023emory,rsna-breast-cancer-detection,pham2022vindr}, outperforming all baselines in multiple downstream tasks.

%% file: floats/fig_teaser.tex
\begin{figure}[ht]
    \centering
    \includegraphics[width=0.85\textwidth]{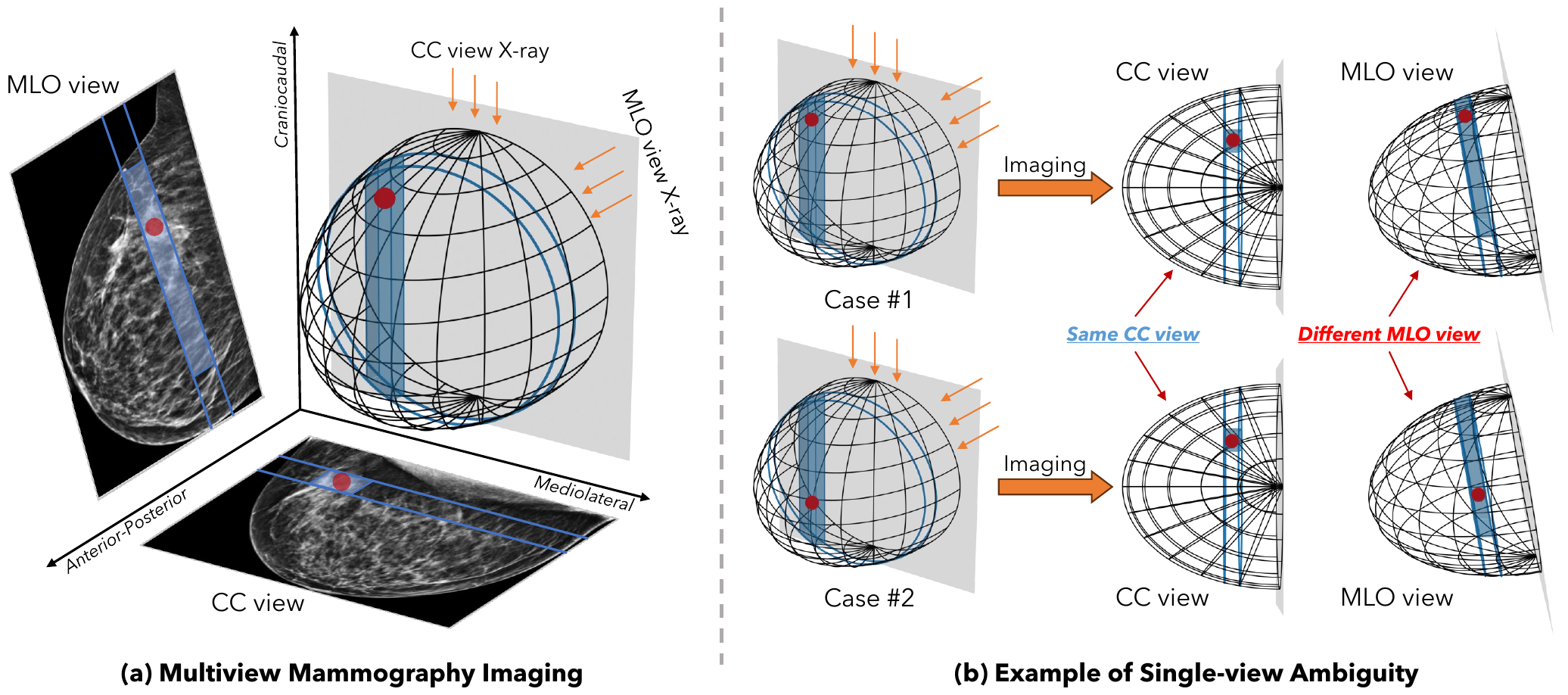}
    \caption{{\bf The Importance of Multi-view.} (a) Due to the imaging process, an ROI (red dot) appears in the same anterior-posterior (AP) slice in both mammography views. (b) Two ROIs located at different positions in the same AP slice result in the same CC view image but different MLO view images, demonstrating the single-view ambiguity.}
    \label{fig:teaser}
\end{figure}

%% file: parts/2_methods.tex
\mysection{Methods}

\input{floats/fig_method}

Given a multi-modal and multi-view mammography dataset $\mathcal{D}=\{(x^{cc}_i, x^{mlo}_i, y_i)\}$, where $i=1,\dots, N$, $(x^{cc}_i, x^{mlo}_i)$ is the multi-view image pair and $y_i$ is the radiology report. Our goal is to learn a robust mammography encoder $f_{V}$ with both global multi-modal knowledge and local multi-view correspondence awareness~(\cref{fig:method}). 

\noindent\textbf{Pre-processing.} Since the MLO view imaging is not parallel to the CC-ML plane~(\cref{fig:teaser}), the mammogram in the MLO view is inclined and contains a pectoral region. We remove the pectoral region using the Hough detector and rotate the image so that the segment between the chest and nipple is parallel to the AP axis, which better aligns the CC and MLO view along the AP axis. However, it is still possible that the two views are misaligned in the AP axis due to extreme cases such as a large pectoral region. We apply a random affine transformation to provide a soft alignment so that the model is more robust to local misalignment. Lastly, we synthesize the radiology report from tabular data following~\cite{du2024multi}, which provides informative structured mammography reports, including imaging information, patient data, and findings. Random text augmentation following~\cite{you2023cxr} is used to generate more diverse reports.

\noindent\textbf{Contrastive Loss.} We first define the contrastive loss $\mathcal{L}$ between two batched embeddings $z$ and $\tilde{z}$ of size $B$ in~\cref{eq:infonce}, following InfoNCE~\cite{chen2020simple} loss: 
\begin{equation}
    \mathcal{L}(z, \tilde{z}) = -\frac{1}{B}\sum^B_{i=1}\log\frac{\exp(\similarity{z_i,~\tilde{z}_i} / \tau)}{\sum^B_{j=1} \exp(\similarity{z_i,~\tilde{z}_j} / \tau)},
    \label{eq:infonce}
\end{equation}
where $\similarity{\cdot,\cdot}$ is the cosine similarity and $\tau$ is the learnable temperature constant. All our learning objectives follow this basic contrastive form.

\mysubsection{Global Multi-view Visual Language Pre-training}

We first conduct multi-view visual language pre-training (VLP) at a global level. We extract the visual feature $(v^{cc}, v^{mlo})$ and textual feature $t$ with corresponding modality encoder $f_V$ and $f_T$. We use the embedding of \texttt{[CLS]} token as the global feature and optimize the multi-view contrastive loss $\mathcal{L}(v^{cc}, v^{mlo})$. Since the multi-view mammograms are different projections of the same breast, representing different views of the same information, it is natural to optimize the image-to-text contrastive loss symmetrically. Thus, our final global optimization objective is:
\begin{equation}
    \mathcal{L}_{global} = \mathcal{L}(v^{cc}, v^{mlo}) + \frac{1}{2}\big[\mathcal{L}(v^{cc}, t) + \mathcal{L}(t, v^{cc}) + \mathcal{L}(v^{mlo}, t) + \mathcal{L}(t, v^{mlo})\big],
    \label{eq:global}
\end{equation}
which optimizes both global multi-view contrastive loss and symmetric image-to-text losses. The textual supervision signal can help the model to learn an embedding space with high-level semantic information.

\mysubsection{Geometry-Guided Local Alignment}
\label{sec:geometric_alignment}

\noindent\textbf{Spatial Attention Aggregation.} We use the patch features from $f_V$ to conduct local alignment. Instead of using raw patch tokens that have a small receptive field, we aggregate the patch features using a spatial attention pooling layer to form $M$ super-patches with a larger receptive field. These super-patches contain higher-level semantic information, which will be used for local alignment.

\noindent\textbf{AP Slice Sampling and Local Alignment.} We use the known geometry of mammography imaging as guidance to align the local patches. Namely, the image slices from both views in the same AP position represent the same tissue in the 3D breast, and each patch in the CC/MLO view should be aligned with a complete slice in the other view.
Thus, we conduct patch-to-slice alignment along the AP axis. For a query patch $q^{cc}_{i,j}$ in $i^{th}$ row and $j^{th}$ column, its corresponding AP slice in the MLO view is $s^{mlo}_j=\{q^{mlo}_{1,j},\dots,q^{mlo}_{\sqrt{M},j}\}$. A multi-head cross-attention module is employed to model the alignment process between $q^{cc}_{i,j}$ and $s^{mlo}_j$,
\begin{equation}
    p^{cc}_{i,j} = \text{CrossAttn.}\big(q^{cc}_{i,j},~s^{mlo}_j,~s^{mlo}_j\big) = \text{softmax}\big(\similarity{q^{cc}_{i,j},~s^{mlo}_j} / \sqrt{d}\big)\cdot s^{mlo}_j,
    \label{eq:crossattn}
\end{equation}
where $d$ is the embedding dimension.  We omit the linear projector for simplicity. The output $p^{cc}_{i,j}$ can be viewed as the weighted sum over the slice $s^{mlo}_j$ based on its correspondence to the query patch.
So, $p^{cc}_{i,j}$ will naturally be the cross-view positive for query patch $q^{cc}_{i,j}$.
Similar computation is used for the MLO patches. 

\noindent\textbf{Negative Samples.}
To enhance the local positional awareness within the mammograms, we use all other patches from different positions as the negatives, \ie, $\mathcal{S}^{cc}_{position} = \{p^{cc}_{m,n}|(m,n)\neq (i,j)\}$, which provides $M-1$ negative samples.
However, using only $\mathcal{S}_{position}$ as negatives may result in a sub-optimal performance as the model can learn to short-cut via using positional encoding. To address this, we use additional negative patches from the same position of different patients across the batch, \ie, $\mathcal{S}^{cc}_{patient}=\{\tilde{p}^{cc}_{i,j}|\tilde{p}^{cc}_{i,j}\neq p^{cc}_{i,j}\}$ and $\tilde{p}^{cc}$ comes from other patients in the batch. These patches are natural negative samples for the query patch since they are from different patients; this forces the model to focus more on patch features rather than positional encoding. The final negative sample set is $\mathcal{S}^{cc} = \mathcal{S}^{cc}_{position}\cup \mathcal{S}^{cc}_{patient}$, providing $M+B-2$ negative samples. The negative set for the MLO view query patches is built similarly.

\noindent\textbf{Final Losses}. 
We optimize the following local alignment loss symmetrically:
\begin{equation}
    \mathcal{L}_{local} = -\frac{1}{2M}\sum^{\sqrt{M}}_{i,j=1}\sum_{\nu\in\{cc,mlo\}}\log\frac{\exp(\similarity{q^{\nu}_{i,j},~p^{\nu}_{i,j}} / \tau)}{\sum_{p^{\nu}\in \mathcal{S}^{\nu}} \exp(\similarity{q^{\nu}_{i,j},~p^{\nu}} / \tau)}.
    \label{eq:local}
\end{equation}
$\mathcal{L}_{local}$ forces the model to align each super-patch with its corresponding AP slice from the other view and ensures the model learns both relative positional relationships and semantic correspondence across both views. The final optimization goal is the sum of global and local loss: $\mathcal{L}_{final} = \mathcal{L}_{global} + \mathcal{L}_{local}$.

%% file: floats/fig_method.tex
\begin{figure}[t]
    \centering
    
    \includegraphics[width=1.0\textwidth]{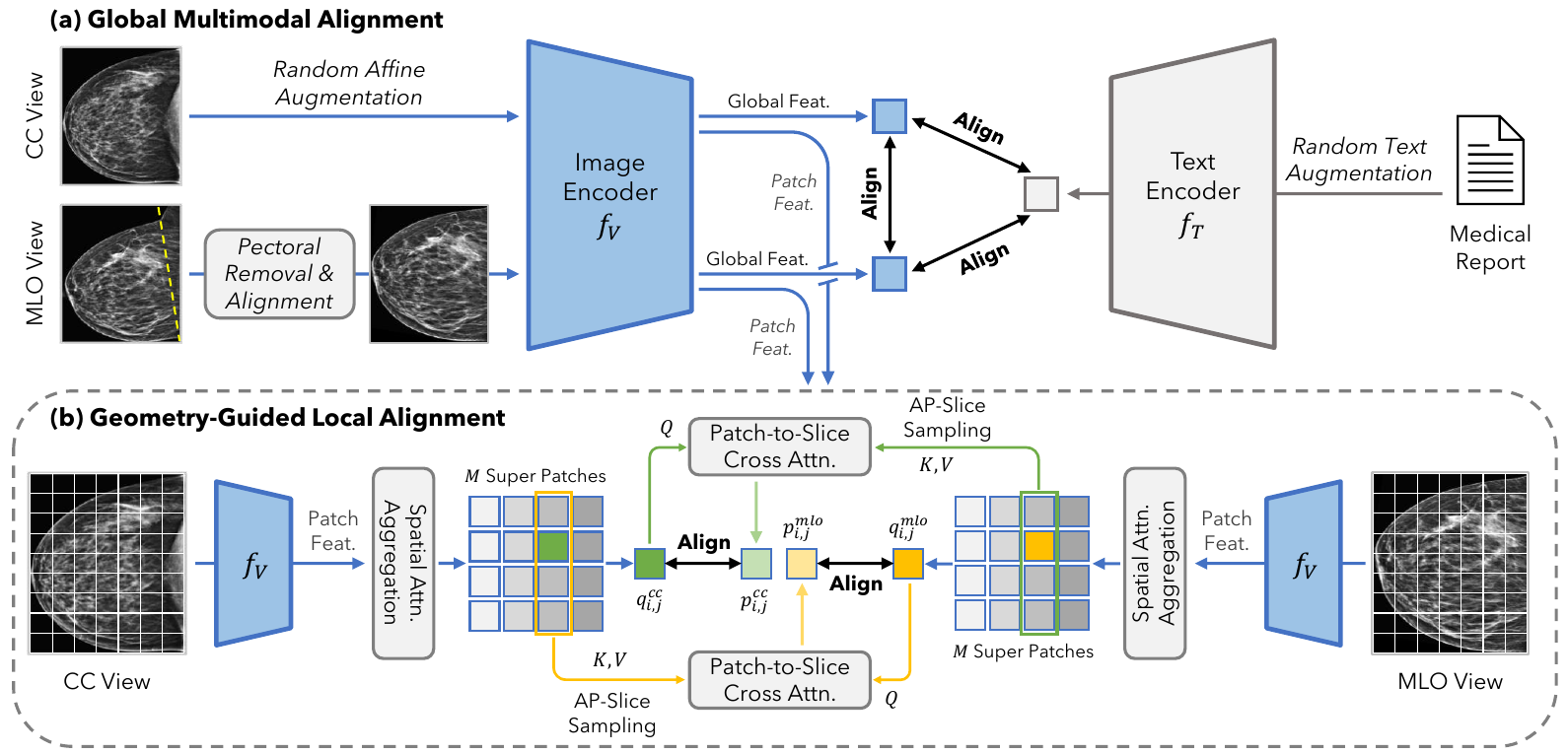}
    \caption{{\bf GLAM Model.}  
    (a) Our method conducts global multi-view CLIP and aligns global visual features to text features from the report. (b) The patch level feature from each view is used to conduct geometry-guided local alignment, where patches from the same AP slice are used as positive matches in a cross-attention mechanism. }
    \label{fig:method}
    \vspace{-2mm}
\end{figure}

%% file: parts/3_experiment.tex
\mysection{Experiments}
\label{sec:experiment}

\input{floats/tab_embed_birads}

\mysubsection{Experimental Settings} 

\noindent\textbf{Datasets.} We pre-train our model on the \textbf{EMBED}~\cite{jeong2023emory} dataset with over 257k screening mammograms with tabular annotated data. We create training/validation/test sets with 70\%/10\%/20\% data, respectively. We evaluate on this dataset for screening BI-RADS (3 classes) and density (4 classes) prediction. We also evaluate on the \textbf{VinDr}~\cite{pham2022vindr} dataset with 20k images for BI-RADS (5 classes) and density (4 classes) prediction using the given data splits. This dataset (from Vietnam) has a different distribution than our pre-training data (from USA). 
We also evaluate on the \textbf{RSNA-Mammo}~\cite{rsna-breast-cancer-detection} dataset for binary cancer prediction. From the provided dataset of 54k images, we split 15\% as the test set. 

\noindent\textbf{Tasks.} We focus on three classification settings for individual mammograms: \emph{zero-shot} on in-domain data, \emph{linear-probing}, and \emph{full fine-tune}. We further vary the size of training data in linear probing to evaluate the data efficiency.

\noindent\textbf{Implementations.} We initialize our encoders using BioClinical-BERT~\cite{alsentzer-etal-2019-publicly} and DiNOv2~\cite{oquab2023dinov2} ViT-B~\cite{dosovitskiy2020image}. We use a batch size of 144, learning rate of $4\times10^{-5}$, and weight decay of $0.2$ to pre-train our model using SGD and cosine learning rate scheduler for 40k steps. For downstream linear probing and fine-tuning, we set batch size to 96, learning rate to $5\times10^{-4}$, and weight decay to $0.001$ and train for 8k steps using SGD. We train with balanced sampling. The same setting is applied to all baselines. All images are resized to $518\times 518$ as input.

\noindent\textbf{Baselines and Metrics.} We compare with vision-only transformers with or without ImageNet~\cite{deng2009imagenet} pre-training; CLIP~\cite{radford2021learning} and SLIP~\cite{mu2022slip} as natural image domain baselines; ConVIRT~\cite{zhang2022contrastive} and MGCA~\cite{wang2022multi} as medical CLIP baselines from different imaging domains; and  Mammo-CLIP~\cite{ghosh2024mammo} and MaMA~\cite{du2024multi} as in-domain baselines. \emph{All baselines except Mammo-CLIP are pre-trained on EMBED}~\cite{jeong2023emory}, just like our model. We use the official pre-trained weights for Mammo-CLIP to show the influence of different pre-training data. We report balanced accuracy (bACC) and AUC as our metrics since the distribution of mammography data is extremely imbalanced; simple accuracy may be biased toward majority classes.

\input{floats/tab_embed_density}

\mysubsection{Results}

\noindent\textbf{In Domain Analysis.} We first evaluate on the in-domain EMBED test set for BI-RADS and density prediction (\cref{tab:birads} and \cref{tab:density}). Our model outperforms all the baselines consistently in all scenarios, surpassing the best baselines by 2.3\% in AUC on average. 
Even with only 1\% of training data, our pre-trained model can still outperform almost all baselines trained with 100\% of data. 
Vision-only methods generally underperform models with VLP.
We note that Mammo-CLIP~\cite{ghosh2024mammo}, pre-trained on $\sim$$10\times$ less data, is 8\% lower on average in bACC, showing a worse generalization capability and highlighting the necessity of scaling the training data. 
Meanwhile, other baselines~\cite{du2024multi,wang2022multi,mu2022slip} that have only global multi-view alignment failed to beat our model since they lack fine-grained multi-view awareness, resulting in suboptimal embedding space.

\input{floats/tab_vindr_rsna}

\noindent\textbf{Out of Domain Analysis.} We further evaluate performance on the out-of-domain datasets VinDr and RSNA-Mammo (\cref{tab:vindr_rsna}) to illustrate the generalization capability of each model. Our model performs the best in 10 out of 12 metrics, suggesting good generalization on unseen data.
We note that the gap between our method and other baselines is smaller in full fine-tune settings compared with linear probing. This is mainly because these out-of-domain datasets have a smaller training set, which makes it easier for the model to converge.

\input{floats/tab_multiview}

\noindent\textbf{Multi-view Analysis.} We evaluate the capability of modeling multi-view correspondence under zero-shot settings, which focus on pre-trained embedding quality (\cref{tab:multiview}). We sub-sampled a test set from EMBED with 7,676 paired multi-view mammograms and evaluated under single- and multi-view prediction, where the multi-view prediction is obtained by averaging the single-view results. Our model improves by $\sim$$2.5\%$ in BI-RADS prediction after switching to multi-view settings, while the baselines have less to no improvement.
Since the density is similar in both views, there is less improvement.
This indicates that our method can model the multi-view geometry and extract complementary features for each view.

\input{floats/tab_ablation}
\input{floats/fig_visualization}

\noindent\textbf{Ablation Study.} Model ablation results are in \cref{tab:ablation}. First, removing the geometry-guided local alignment learning greatly harms the model's performance, especially its robustness under full fine-tuning. Same-position negatives provide $\sim$2\% improvement in zero-shot and linear probing and ensure stable behavior in full fine-tuning. Replacing the spatial attention aggregation with average pooling also results in sub-optimal performance. Lastly, we evaluate the necessity of following the geometry guidance in local correspondence learning by computing the attention across all patches in the view. This lowers performance since the geometry constraint is broken. We also test different numbers of super-patches $M$, where patch size will influence the performance as discussed in \cref{sec:geometric_alignment}.

\noindent\textbf{Qualitative Visualization.} We visualize the cross-view patch-to-slice attention weights in \cref{fig:vis}. We pick random patches within annotated ROIs from the EMBED test set and visualize their attention scores in the corresponding AP slice in the other view. 
Our model can accurately locate the ROI in the other view and, therefore, gain multi-view awareness during pre-training.

%% file: floats/tab_embed_birads.tex
\begin{table}[!t]
\centering
\caption{\textbf{BI-RADS Prediction Results on EMBED.} Performance (in \%) for each method under zero-shot, linear probing with varying training data size, and full fine-tune settings. $^*$ denotes use of official pre-trained weights. Best and second-best results are in bold and underlined, respectively. Our method is shaded in gray.}
\label{tab:birads}
\setlength{\tabcolsep}{8pt}
\resizebox{\textwidth}{!}
{
\begin{tabular}{lcccccccccc}
\toprule
\multicolumn{1}{c}{\multirow{3}{*}{\textbf{Methods}}} & \multicolumn{2}{c}{\textbf{Zero-shot}} & \multicolumn{6}{c}{\textbf{Linear Probing}} & \multicolumn{2}{c}{\textbf{Full Fine-tune}} \\ \cmidrule(l){2-3} \cmidrule(l){4-9} \cmidrule(l){10-11}
& \multicolumn{2}{c}{\textbf{100\%}} & \multicolumn{2}{c}{\textbf{1\%}} & \multicolumn{2}{c}{\textbf{10\%}} & \multicolumn{2}{c}{\textbf{100\%}} & \multicolumn{2}{c}{\textbf{100\%}} \\
 & bACC & AUC & bACC & AUC & bACC & AUC & bACC & AUC & bACC & AUC \\ \midrule\midrule
\textit{Vision only} &  &  &  &  &  &  &  &  &  &  \\
~~Random-ViT~\cite{dosovitskiy2020image} & - & - & 35.19 & 52.56 & 36.36 & 52.79 & 36.05 & 52.76 & 35.73 & 52.42 \\
~~DiNOv2-ViT~\cite{oquab2023dinov2} & - & - & 41.48 & 57.62 & 45.97 & 61.64 & 45.45 & 61.53 & 43.46 & 60.33 \\ \midrule
\textit{CLIP pre-trained} &  &  &  &  &  &  &  &  &  &  \\
~~CLIP~\cite{radford2021learning} & 37.17 & 55.90 & 43.35 & 59.57 & 47.89 & 64.46 & 47.05 & 63.50 & 45.77 & 61.79 \\
~~SLIP~\cite{mu2022slip} & 44.24 & 60.67 & 43.43 & 60.39 & 48.82 & 64.67 & 46.66 & 63.35 & 37.81 & 54.60 \\
~~ConVIRT~\cite{zhang2022contrastive} & 43.02 & 61.31 & 47.45 & 63.16 & 47.91 & 63.78 & 47.73 & 63.40 & 49.41 & 65.41 \\
~~MGCA~\cite{wang2022multi} & {\ul 45.48} & {\ul 61.92} & {\ul 47.81} & 62.76 & 48.30 & 63.44 & {\ul 48.82} & {\ul 64.83} & {\ul 50.37} & 65.70 \\
~~Mammo-CLIP-B2$^*$~\cite{ghosh2024mammo} & 36.93 & 56.20 & 42.05 & 60.67 & 42.90 & 61.80 & 42.53 & 62.18 & 43.03 & 60.75 \\
~~Mammo-CLIP-B5$^*$~\cite{ghosh2024mammo} & 36.67 & 57.09 & 38.68 & 59.93 & 38.15 & 61.21 & 38.29 & 61.25 & 38.58 & 61.46 \\
~~MaMA ~\cite{du2024multi} & 44.61 & 61.63 & 46.63 & {\ul 63.65} & {\ul 48.90} & {\ul 64.81} & 47.96 & 63.69 & 49.95 & {\ul 66.06} \\ \midrule \rowcolor[HTML]{EFEFEF} 
~~GLAM (Ours) & \textbf{47.24} & \textbf{64.86} & \textbf{48.57} & \textbf{64.71} & \textbf{49.17} & \textbf{65.29} & \textbf{50.07} & \textbf{66.59} & \textbf{51.81} & \textbf{67.34} \\ \bottomrule
\end{tabular}
}

\end{table}

%% file: floats/tab_embed_density.tex
\begin{table}[!t]
\centering
\caption{\textbf{Density Prediction Results on EMBED.} Performance (in \%)   for each method under zero-shot, linear-probing with varying training data size, and full fine-tune settings. $^*$ denotes use of official pre-trained weights. Best and second-best results are in bold and underlined, respectively. Our method is shaded in gray.}
\label{tab:density}
\setlength{\tabcolsep}{8pt}
\resizebox{\textwidth}{!}
{
\begin{tabular}{lcccccccccc}
\toprule
\multicolumn{1}{c}{\multirow{3}{*}{\textbf{Methods}}} & \multicolumn{2}{c}{\textbf{Zero-shot}} & \multicolumn{6}{c}{\textbf{Linear Probing}} & \multicolumn{2}{c}{\textbf{Full Fine-tune}} \\ \cmidrule(l){2-3} \cmidrule(l){4-9} \cmidrule(l){10-11}
& \multicolumn{2}{c}{\textbf{100\%}} & \multicolumn{2}{c}{\textbf{1\%}} & \multicolumn{2}{c}{\textbf{10\%}} & \multicolumn{2}{c}{\textbf{100\%}} & \multicolumn{2}{c}{\textbf{100\%}} \\
 & bACC & AUC & bACC & AUC & bACC & AUC & bACC & AUC & bACC & AUC \\ \midrule\midrule
\textit{Vision only} &  &  &  &  &  &  &  &  &  &  \\
~~Random-ViT~\cite{dosovitskiy2020image} & - & - & 38.99 & 68.99 & 41.35 & 68.91 & 41.48 & 69.03 & 64.55 & 86.44 \\
~~DiNOv2-ViT~\cite{oquab2023dinov2} & - & - & 65.62 & 87.06 & 67.54 & 87.39 & 67.54 & 87.45 & 77.47 & 93.40 \\ \midrule
\textit{CLIP pre-trained} &  &  &  &  &  &  &  &  &  &  \\
~~CLIP~\cite{radford2021learning} & 59.69 & 88.73 & 74.77 & {\ul 93.32} & 76.15 & 91.83 & 76.92 & 92.91 & 78.32 & {\ul 93.91} \\
~~SLIP~\cite{mu2022slip} & {\ul 78.06} & {\ul 92.78} & 75.47 & 93.19 & {\ul 77.22} & 92.90 & {\ul 77.99} & {\ul 93.75} & {\ul 78.81} & 93.89 \\
~~ConVIRT~\cite{zhang2022contrastive} & 61.48 & 72.54 & 73.64 & 92.77 & 74.20 & 92.16 & 75.27 & 92.70 & 78.31 & 93.85 \\
~~MGCA~\cite{wang2022multi} & 62.45 & 71.29 & 72.34 & 91.48 & 72.89 & 91.92 & 73.56 & 91.98 & 78.37 & 93.66 \\
~~Mammo-CLIP-B2$^*$~\cite{ghosh2024mammo} & 53.50 & 80.50 & 70.02 & 88.91 & 68.98 & 88.59 & 69.22 & 88.81 & 76.01 & 92.47 \\
~~Mammo-CLIP-B5$^*$~\cite{ghosh2024mammo} & 46.07 & 71.89 & 69.60 & 89.47 & 70.23 & 89.98 & 69.46 & 89.96 & 69.90 & 90.05 \\
~~MaMA ~\cite{du2024multi} & 75.18 & 91.81 & {\ul 74.88} & 92.79 & 76.74 & {\ul 93.15} & 73.67 & 91.69 & 77.61 & 92.66 \\ \midrule \rowcolor[HTML]{EFEFEF} 
~~GLAM (Ours) & \textbf{79.06} & \textbf{93.76} & \textbf{77.87} & \textbf{93.65} & \textbf{78.76} & \textbf{94.01} & \textbf{79.61} & \textbf{94.03} & \textbf{80.32} & \textbf{94.05} \\ \bottomrule
\end{tabular}
}

\end{table}

%% file: floats/tab_vindr_rsna.tex
\begin{table}[!t]
\centering
\caption{\textbf{Results on VinDr and RSNA-Mammo.} Performance (in \%) for each method on prediction tasks on VinDr and RSNA-Mammo datasets under linear-probing and full fine-tune settings. $^*$ denotes use of official pre-trained weights. Best and second-best results are in bold and underlined, respectively. Our method is shaded in gray.}
\label{tab:vindr_rsna}
\setlength{\tabcolsep}{8pt}
\resizebox{\textwidth}{!}
{
\begin{tabular}{lcccccccccccc}
\toprule
\multicolumn{1}{c}{\multirow{3}{*}{\textbf{Methods}}} & \multicolumn{4}{c}{\textbf{VinDr - BI-RADS}} & \multicolumn{4}{c}{\textbf{VinDr - Density}} & \multicolumn{4}{c}{\textbf{RSNA - Cancer}} \\ \cmidrule(l){2-5} \cmidrule(l){6-9} \cmidrule(l){10-13} 
\multicolumn{1}{c}{} & \multicolumn{2}{c}{\textbf{Linear Probing}} & \multicolumn{2}{c}{\textbf{Full Fine-tune}} & \multicolumn{2}{c}{\textbf{Linear Probing}} & \multicolumn{2}{c}{\textbf{Full Fine-tune}} & \multicolumn{2}{c}{\textbf{Linear Probing}} & \multicolumn{2}{c}{\textbf{Full Fine-tune}} \\
\multicolumn{1}{c}{} & bACC & AUC & bACC & AUC & bACC & AUC & bACC & AUC & bACC & AUC & bACC & AUC \\ \midrule
\textit{Vision only} &  &  &  &  &  &  &  &  & \multicolumn{1}{l}{} & \multicolumn{1}{l}{} & \multicolumn{1}{l}{} & \multicolumn{1}{l}{} \\
~~Random-ViT~\cite{dosovitskiy2020image} & 27.28 & 56.50 & 26.58 & 59.62 & 29.69 & 56.95 & 34.25 & 37.46 & 54.75 & 57.06 & 54.18 & 58.31 \\
~~DiNOv2-ViT~\cite{oquab2023dinov2} & 35.47 & 63.51 & 37.99 & 64.36 & 59.68 & 86.38 & 69.49 & 92.96 & 52.94 & 61.01 & 52.25 & 66.57 \\ \midrule
\textit{CLIP pre-trained} &  &  &  &  &  &  &  &  & \multicolumn{1}{l}{} & \multicolumn{1}{l}{} & \multicolumn{1}{l}{} & \multicolumn{1}{l}{} \\
~~CLIP~\cite{radford2021learning} & 41.26 & 68.92 & 41.11 & 72.36 & 70.85 & {\ul 93.07} & 72.38 & 92.52 & {\ul 65.79} & {\ul 71.93} & 63.69 & 67.94 \\
~~SLIP~\cite{mu2022slip} & 40.03 & 70.18 & 41.06 & {\ul 74.23} & 71.98 & 92.79 & 66.37 & 85.80 & 60.05 & 65.78 & 55.65 & 61.14 \\
~~ConVIRT~\cite{zhang2022contrastive} & 39.11 & 71.64 & 39.95 & 72.18 & 63.63 & 71.01 & 71.01 & 90.05 & 62.74 & 68.80 & 53.59 & 66.03 \\
~~MGCA~\cite{wang2022multi} & 38.85 & {\ul 72.72} & 40.85 & 73.27 & 71.82 & 89.63 & 76.94 & 90.51 & 64.64 & 69.94 & 54.69 & 68.46 \\
~~Mammo-CLIP-B2$^*$~\cite{ghosh2024mammo} & 34.68 & 64.76 & 36.23 & 65.92 & 64.09 & 87.90 & 64.26 & 87.91 & 52.81 & 61.52 & 53.55 & 61.32 \\
~~Mammo-CLIP-B5$^*$~\cite{ghosh2024mammo} & 39.68 & 67.58 & \textbf{42.78} & 71.83 & 70.70 & 87.64 & \textbf{78.56} & {\ul 93.28} & 60.72 & 66.02 & {\ul 64.50} & {\ul 72.96} \\
~~MaMA ~\cite{du2024multi} & {\ul 41.35} & 68.36 & 35.94 & 61.78 & {\ul 73.49} & 92.77 & 65.63 & 92.64 & 63.18 & 69.32 & 57.31 & 62.28 \\ \midrule \rowcolor[HTML]{EFEFEF} 
~~GLAM (Ours) & \textbf{41.41} & \textbf{73.81} & {\ul 41.87} & \textbf{74.82} & \textbf{74.58} & \textbf{93.60} & {\ul 78.27} & \textbf{93.94} & \textbf{67.45} & \textbf{73.14} & \textbf{68.77} & \textbf{75.04} \\ \bottomrule
\end{tabular}
}

\end{table}

%% file: floats/tab_multiview.tex
\begin{table}[!t]
\centering
\caption{\textbf{Multi-view Prediction Results.}  Zero-shot performance (in \%) of BI-RADS and density prediction under single-view and multi-view settings. $^*$ denotes use of official pre-trained weights. Best results are in bold. Our method is shaded in gray.}
\label{tab:multiview}
\setlength{\tabcolsep}{8pt}
\resizebox{0.85\textwidth}{!}
{
\begin{tabular}{lcccccccc}
\toprule
\multicolumn{1}{c}{\multirow{3}{*}{\textbf{Methods}}} & \multicolumn{4}{c}{\textbf{EMBED - BI-RADS}} & \multicolumn{4}{c}{\textbf{EMBED - Density}} \\ \cmidrule(l){2-5} \cmidrule(l){6-9}
\multicolumn{1}{c}{} & \multicolumn{2}{c}{\textbf{Single-view}} & \multicolumn{2}{c}{\textbf{Multi-view}} & \multicolumn{2}{c}{\textbf{Single-view}} & \multicolumn{2}{c}{\textbf{Multi-view}} \\
\multicolumn{1}{c}{} & bACC & AUC & bACC & AUC & bACC & AUC & bACC & AUC \\ \midrule
~~CLIP~\cite{radford2021learning} & 42.35 & 61.79 & 44.47 & 63.56 & 57.22 & 87.40 & 57.65 & 88.08 \\
~~MGCA~\cite{wang2022multi} & 44.72 & 62.17 & 46.17 & 63.23 & 68.49 & 89.87 & 71.23 & 91.30 \\
~~Mammo-CLIP-B2$^*$~\cite{ghosh2024mammo} & 36.80 & 56.70 & 36.36 & 56.98 & 53.29 & 80.24 & 54.65 & 81.04 \\
~~Mammo-CLIP-B5$^*$~\cite{ghosh2024mammo} & 38.27 & 58.35 & 38.30 & 58.64 & 49.63 & 72.63 & 50.20 & 73.42 \\ \midrule \rowcolor[HTML]{EFEFEF} 
~~GLAM (Ours) & \textbf{46.05} & \textbf{63.28} & \textbf{48.40} & \textbf{66.02} & \textbf{79.02} & \textbf{93.63} & \textbf{79.42} & \textbf{94.08} \\ \bottomrule
\end{tabular}
}

\end{table}

%% file: floats/tab_ablation.tex
\begin{table}[!t]
\centering
\caption{\textbf{Ablation Results.} Performance (in \%) of BI-RADS prediction on EMBED for each ablated model. GLA: Geometry-guided Local Alignment; SPN: Same Position Negatives; SAA: Spatial Attention Aggregation. Best results are in bold. Our method is shaded in gray.}
\label{tab:ablation}
\setlength{\tabcolsep}{8pt}
\resizebox{0.93\textwidth}{!}
{
\begin{tabular}{ccccccccccccc}
\toprule
\multirow{2}{*}{\textbf{GLA}} & \multirow{2}{*}{\textbf{SPN}} & \multirow{2}{*}{\textbf{SAA}} & \multirow{2}{*}{\textbf{AP Sampling}} & \multicolumn{3}{c}{\textbf{\#Local Regions}} & \multicolumn{2}{c}{\textbf{Zero-shot}} & \multicolumn{2}{c}{\textbf{Linear Probing}} & \multicolumn{2}{c}{\textbf{Full Fine-tune}} \\ \cmidrule(l){5-7} \cmidrule(l){8-9} \cmidrule(l){10-11} \cmidrule(l){12-13} 
 &  &  &  & $M=16$ & $M=81$ & $M=324$ & bACC & AUC & bACC & AUC & bACC & AUC \\ \midrule
 &  &  &  &  &  &  & 45.12 & 62.82 & 49.36 & 65.33 & 37.81 & 54.60 \\
\checkmark &  & \checkmark & \checkmark &  & \checkmark &  & 45.55 & 62.23 & 47.48 & 63.75 & 36.34 & 53.85 \\
\checkmark & \checkmark &  & \checkmark &  & \checkmark &  & 44.24 & 60.67 & 46.66 & 63.35 & 37.68 & 53.99 \\
\checkmark & \checkmark & \checkmark &  &  & \checkmark &  & 44.98 & 62.56 & 49.17 & 65.24 & 48.45 & 64.45 \\
\checkmark & \checkmark & \checkmark & \checkmark & \checkmark &  &  & 43.75 & 60.81 & 46.94 & 64.03 & 46.80 & 63.51 \\ 
\checkmark & \checkmark & \checkmark & \checkmark &  &  & \checkmark & 45.99 & 62.72 & 48.55 & 64.76 & 47.58 & 63.07 \\ \midrule \rowcolor[HTML]{EFEFEF} 
\checkmark & \checkmark & \checkmark & \checkmark &   & \checkmark &  & \textbf{47.24} & \textbf{64.86} & \textbf{50.07} & \textbf{66.59} & \textbf{51.81} & \textbf{67.34} \\ \bottomrule
\end{tabular}
}

\end{table}

%% file: floats/fig_visualization.tex
\begin{figure}[t]
    \centering
    \includegraphics[width=0.92\textwidth]{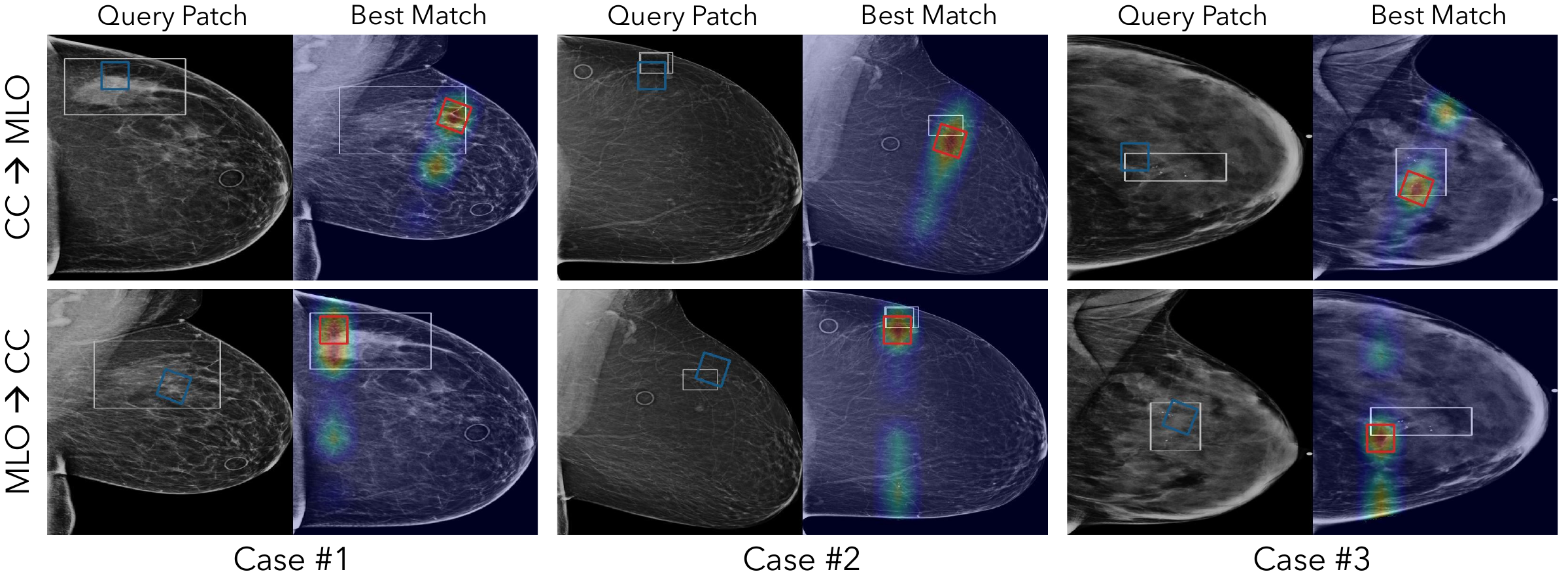}
    \caption{{\bf Cross View Patch-to-Slice Attention Visualization}
    For each pair of mammograms, the white bounding boxes indicate the ROIs, \eg, tumor; the blue box is the query patch; and the red box is the patch with the highest attention. Patches in the MLO view are inclined due to AP alignment during pre-processing.}
    \label{fig:vis}
\end{figure}

%% file: parts/4_conclusion.tex
\mysection{Discussion and Conclusion}

We proposed one of the largest screening mammography foundation CLIP models to date, \ie, GLAM, with a novel geometry-guided local alignment module to enable the fine-grained cross-view awareness of the model. The proposed method achieved state-of-the-art performance in three different datasets compared with existing VLP models. While we mainly focus on evaluating the quality of the pre-trained embedding space, we also plan to fuse our robust backbone with multi-view fusion methods to further improve the performance and clinical applicability. Future plans include introducing dense multi-modal contrastive learning and extending multi-view alignment to both sides of the breast. 

\subsubsection{Acknowledgments} This work was supported by NIH grant R21EB032950.

\subsubsection{Disclosure of Interests} The authors have no competing interests in this work and other related research.